\pdfoutput=1

\documentclass[11pt]{article}
\usepackage{acl}
\usepackage{times}
\usepackage{latexsym}

\usepackage{colortbl}
\usepackage{xcolor}
\usepackage[T1]{fontenc}
\usepackage[utf8]{inputenc}
\usepackage{microtype}
\usepackage{inconsolata}

\usepackage{amsthm}

\newtheorem{observation}{Observation}

\usepackage{amsmath}
\usepackage{graphicx}
\usepackage{algorithm}
\usepackage[noend]{algpseudocode}

\usepackage{enumitem}

\usepackage{multirow}
\usepackage{adjustbox}
\usepackage{bbm}
\usepackage{wrapfig,lipsum}
\usepackage{xspace}
\usepackage{booktabs,cellspace}  
\usepackage[normalem]{ulem}
\usepackage{makecell}
\usepackage{amssymb}
\usepackage{pifont}
\usepackage{todonotes}
\usepackage{lipsum}
\usepackage{afterpage}

\usepackage{microtype}

\usepackage{cleveref}

\newcommand{\llamaM}{\textsc{LLaVA-v1.6-Mistral-7B}\xspace}

\newcommand{\llamaT}{\textsc{Llama-3.1-8B-Instruct}\xspace}
\newcommand{\PhiV}{\textsc{Phi-3.5-Vision-Instruct}\xspace}

\newcommand{\qwenA}{\textsc{Qwen2-VL-7B-Instruct}\xspace}
\newcommand{\qwenB}{\textsc{Qwen2.5-VL-7B-Instruct}\xspace}
\newcommand{\gpt}{\textsc{GPT-4o}\xspace}
\newcommand{\latex}{\textsc{LaTeX}\xspace}

\newcommand\blfootnote[1]{%
  \begingroup
  \renewcommand\thefootnote{}\footnote{#1}%
  \addtocounter{footnote}{-1}%
  \endgroup
}

\usepackage{etoolbox}
\usepackage[tikz]{bclogo}
\usepackage{subcaption}
\usepackage{adjustbox}

\definecolor{msftBlue}{RGB}{0,164,239}
\definecolor{msftGreen}{RGB}{127,186,0}
\definecolor{msftYello}{RGB}{255,185,0}
\definecolor{msftBlack}{RGB}{0,0,0}

\usepackage{tcolorbox} 
\tcbuselibrary{skins} 
\usepackage[T1]{fontenc}

\usepackage{tcolorbox} 
\tcbuselibrary{skins} 
\usepackage[T1]{fontenc}

\tcbset{
    userstyle/.style={
        enhanced,
        colback=white,
        colframe=black,
        colbacktitle=gray!20,
        coltitle=black,
        rounded corners,
        sharp corners=north,
        boxrule=0.5pt,
        drop shadow=black!50!white,
        attach boxed title to top left={
            xshift=-2mm,
            yshift=-2mm
        },
        boxed title style={
            rounded corners,
            size=small,
            colback=gray!20
        }
    },
    replystyleg/.style={
        enhanced,
        colback=green!15,
        colframe=black,
        colbacktitle=green!30,
        coltitle=black,
        boxrule=0.5pt,
        drop shadow=black!50!white,
        rounded corners,
        sharp corners=north,
        attach boxed title to top right={
            xshift=-2mm,
            yshift=-2mm
        },
        boxed title style={
            rounded corners,
            size=small,
            colback=green!40
        }
    },
    replystyler/.style={
        enhanced,
        colback=red!15,
        colframe=black,
        colbacktitle=red!40,
        coltitle=black,
        boxrule=0.5pt,
        drop shadow=black!50!white,
        rounded corners,
        sharp corners=north,
        attach boxed title to top right={
            xshift=-2mm,
            yshift=-2mm
        },
        boxed title style={
            rounded corners,
            size=small,
            colback=red!40
        }
    }
}
\newtcolorbox{userquery}[1][]{
    userstyle,
    title=Probed Fuses for \gpt,
    #1
}

\title{Structured Attention Matters to Multimodal LLMs in \\ Document Understanding}

\author{
\bf Chang Liu$^{1}$, Hongkai Chen$^{1\dagger}$, Yujun Cai$^2$, Hang Wu$^{1,3}$, \\
\bf Qingwen Ye$^1$, Ming-Hsuan Yang$^3$, Yiwei Wang$^3$ \\
$^1$vivo Mobile Communication Co., Ltd,
$^2$The University of Queensland, \\
$^3$University of California, Merced, \\
\texttt{changliu.hapo@gmail.com, allenhkchen@gmail.com}
}
\date{}

\begin{document}
\maketitle

\blfootnote{$^\dagger$The corresponding author.}
\blfootnote{$^\dagger$The work was done during the first author's internship at vivo company.}

\vspace{0mm}
\begin{abstract}
Document understanding remains a significant challenge for multimodal large language models (MLLMs). While previous research has primarily focused on locating evidence pages through precise multimodal queries, our work investigates a fundamental yet overlooked aspect: how input format influences document comprehension performance. Through systematic analysis, we discover that raw OCR text often impairs rather than improves MLLMs' performance, which is a counterintuitive finding we attribute to attention dispersion and structure loss. To further substantiate our hypothesis, we propose a novel structure-preserving approach that encodes document elements using the \latex paradigm, maintaining the hierarchical organization and spatial relationships critical for comprehension. Our attention analysis reveals that structured text induces structured attention patterns on both textual and visual content, directing models to focus on semantically meaningful regions while reducing attention waste. This approach significantly enhances MLLMs' document question answering performance across diverse document types without requiring architectural modifications or additional training.
\end{abstract}



\begin{figure}[!tb]
    \centering
	\includegraphics[width=1\linewidth]{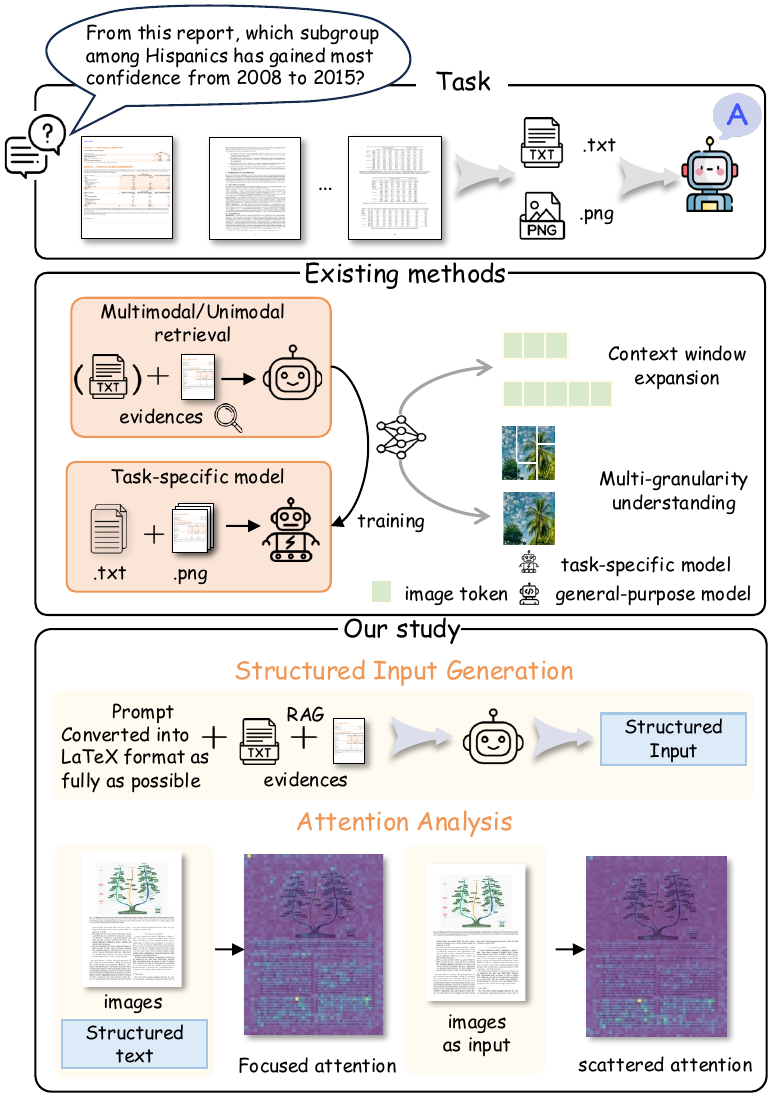}
	\caption{Comparison of Different Approaches for DocQA: Previous research methods focused on using RAG to precisely locate the evidence and then directly input the evidence into general-purpose MLLMs, or on designing task-specific models that focus on multi-granularity extraction of image information and expanding the context window. We propose a novel structure-preserving approach based on the \latex paradigm to explore the impact of input format on general models' responses to DocQA and investigate potential causes through attention analysis.}	\label{fig:teaser}
\end{figure}

\section{Introduction}
Document understanding is a common task in daily life. With the rapid development of multimodal large language models (MLLMs), capable general-purpose systems are increasingly expected to comprehend documents effectively (\cite{zhu2023minigpt,liu2024improved,liu2023visual,wang2024qwen2,Qwen-VL,qwen2.5-VL,anthropicclaudeblog,dong2024internlm,chen2024far,li2023mimic,wang2024cogvlm}). The difficulty in document question answering arises from the diverse nature of questions and the variety of information required, which evidence elements include text blocks, charts, diagrams, and figures, often requiring integration of multiple information sources.
Document understanding through MLLMs presents three key challenges: (1) the information diversity challenge of processing heterogeneous elements, (2) the context integration challenge of synthesizing scattered information, and (3) the structural relationship challenge of understanding how elements relate spatially and logically—relationships intuitive for humans but difficult for machines (\cite{han2025mdocagent}).


Previous research has addressed these challenges primarily by extending context windows to accommodate more content or by developing specialized architectures for extracting multi-granularity information (\cite{ding2022v,hu2024mplug,ye2023mplug,liu2024textmonkey,park2024hierarchical,tito2023hierarchical}). 
With general-purpose MLLMs, the trend has shifted toward retrieval-augmented generation (RAG), which locates relevant evidence and feeds it into models as images, OCR text, or both (\cite{mishra2019ocr,suri2024visdom,park2024hierarchical}). Despite these advances, a critical question remains unexplored: how does the format of input information, rather than merely its content—influence document understanding?  Current approaches typically extract OCR text but discard critical structural information (\cite{khattab2020colbert,faysse2024colpali}). This leads to a counterintuitive phenomenon we observed across multiple datasets: unstructured OCR text often degrades rather than enhances MLLM performance compared to using images alone (\cite{cho2024m3docrag,deng2024longdocurl,han2025mdocagent,zhang2024ocr}). This observation motivated us to investigate the relationship between input structure and model comprehension. We discovered that information structure fundamentally shapes how MLLMs allocate attention across document elements. With unstructured text, models exhibit scattered attention patterns, wasting computational resources and struggling to identify relationships between elements.

Our study includes two main steps. First, we propose an approach to preserve the structure of OCR text. We input the evidence OCR text and images into an MLLM and prompt it to interpret the images with the help of the OCR text as much as possible. This step aims to maintain the structure of diagrams, tables, and texts. The layout information can be preserved in the form of text, which is helpful during the answer generation process. We input the images along with the \latex paradigm to generate answers, exploring the importance of document layout and the structure of sub-elements on the evidence page. We find that the accuracy of MLLMs in answering questions is significantly improved. Second, we analyze the attention distribution and transformation with different inputs, comparing the cases of using images alone and images combined with structured text as input. We find that attention is more focused when constrained by the structured input.

Overall, our findings show that the ability of general-purpose models in document understanding can be improved simply by changing the input format. Furthermore, the attention transformation between different inputs demonstrates that the structured attention brought by structured input is key to improving the ability of MLLMs to answer DocQAs. This indicates that there are other aspects worth exploring beyond focusing solely on enhancing the effectiveness of Retrieval Augmented Generation (RAG).

\begin{figure*}[!tb]
	\centering
	\includegraphics[width=1.0\linewidth]{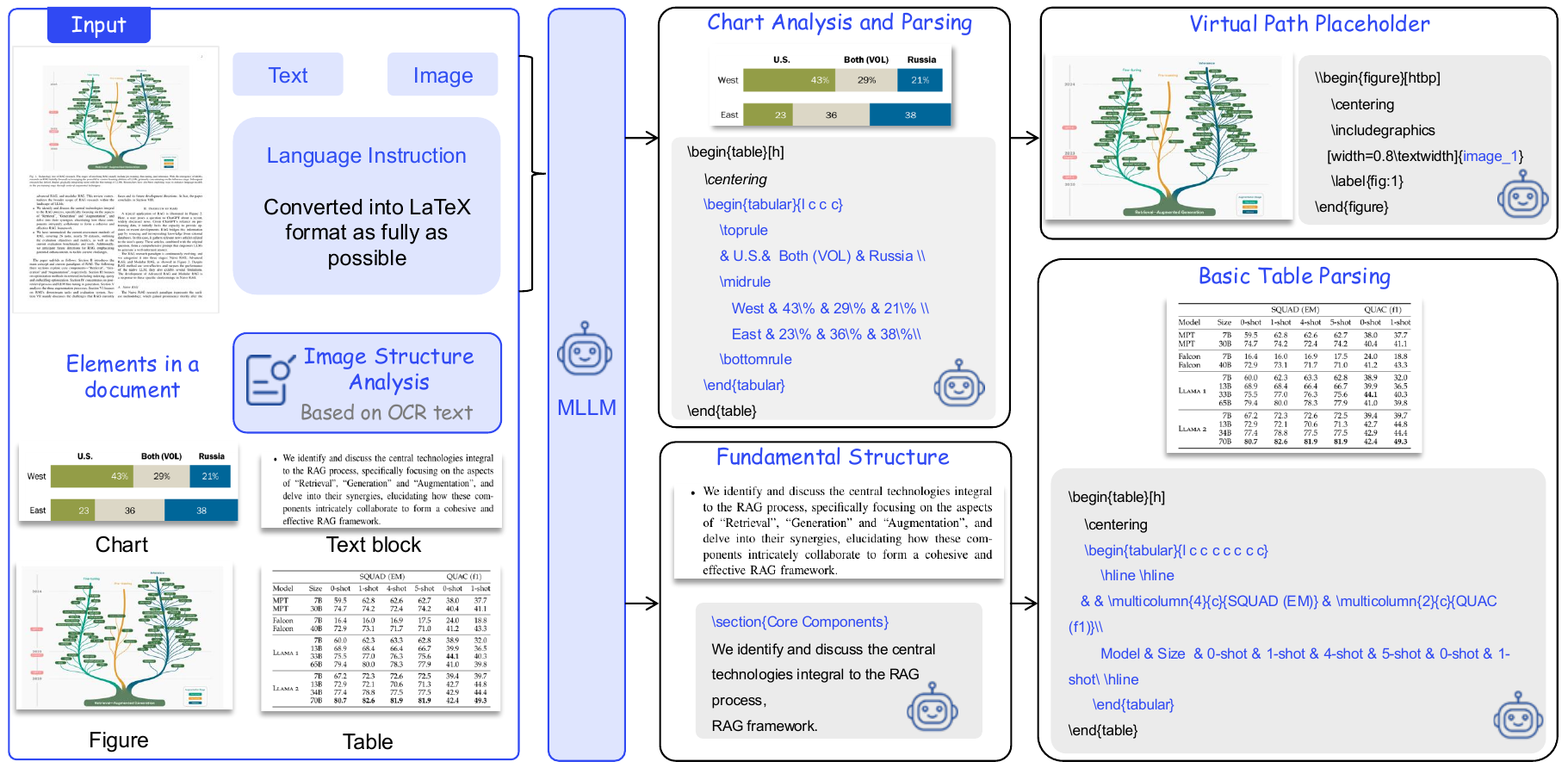}
	\caption{Structured text is generated using \latex. We prompt the MLLM to capture the layout of the given images as accurately as possible, producing blocks that include text, charts, and data tables. Figures that cannot be parsed into concrete content are represented using virtual paths and the \latex paradigm. This approach is simple and only requires API-level access with instruction-level control.}	
    \label{fig:latex element}
\end{figure*}
\vspace{0mm}

\section{Related Work}
We discuss two lines of related work: MLLMs in DocQA tasks and methods used in DocQA tasks.

\paragraph{MLLMs in DocQA Tasks.} 
Recent Document Visual Question Answering (DocVQA) models focus on expanding context window sizes, improving fine-grained visual comprehension, and enhancing layout analysis (\cite{hu2024mplug, liu2024textmonkey, tworkowski2023focused}). These systems often boost performance through architectural enhancements (e.g., added modules) or multi-stage training pipelines with diverse data inputs (\cite{park2024hierarchical, tito2023hierarchical, chen2023longlora}). While general multimodal models like Gemini (\cite{team2024gemini}) and Qwen-VL (\cite{Qwen-VL}) show improved visual processing and context handling, they remain constrained by input length limitations and multi-page processing capabilities. Their reliance on image-derived representations also hinders precise localization of detailed elements, while layout extraction from visual data risks diverting the model's focus, reducing answer accuracy (\cite{han2025mdocagent}).

\paragraph{Methods in DocQA Tasks.}
Retrieval augmentation has become pivotal for adapting general multimodal models to document question answering, addressing input constraints and information overload by retrieving fine-grained evidence through multimodal or unimodal queries (\cite{lewis2020retrieval, gao2023retrieval, chen2024mllm, cho2024m3docrag}). While current methods enhance accuracy through optimized retrieval strategies (\cite{deng2024longdocurl, lu2024text}), input formatting remains challenging: image-only inputs hinder textual comprehension due to incomplete OCR integration, whereas text-only inputs (via OCR) neglect visual elements critical for image-dependent questions. Multimodal inputs combining text and images risk overwhelming models with redundant or conflicting data, leading to attention dispersion. To maximize performance, refining input strategies—such as modality prioritization, structured evidence fusion, or dynamic filtering—is essential for balancing information richness and focused reasoning in retrieved evidence.

\section{Methodology}
Understanding documents through multimodal large language models (MLLMs) represents a complex challenge at the intersection of vision and language processing. Unlike previous approaches that focus primarily on evidence retrieval or context window expansion, our research investigates how input representation fundamentally shapes model comprehension. In this section, we introduce a novel structure-preserving
approach that significantly enhances MLLMs' document understanding capabilities without requiring architectural modifications or additional training. We analyze the attention transformation under different inputs, revealing that structured text induces structured attention patterns on both textual and visual content, enabling the comprehension of MLLMs.


\vspace{0mm}

\subsection{Input Representation in Document Understanding
}\label{sec:3.1}
Document understanding requires models to interpret diverse information types—text blocks, tables, figures, charts, and their interrelationships. Through systematic analysis of MLLM performance in document question answering (DocQA) tasks, we identified a critical paradox


\begin{observation}\label{obs:1}
Providing unstructured OCR text alongside document images often degrades MLLM performance across benchmark datasets, despite increasing the total information available to the model.
\end{observation}

This counterintuitive finding contradicts the common assumption in retrieval-augmented generation that more textual context improves performance. To investigate this phenomenon, we conducted a series of controlled experiments comparing MLLMs' attention patterns when processing: (1) document images alone, (2) images with unstructured OCR text, and (3) images with our proposed structured text representation.


\subsection{OCR text with structure}\label{sec:3.2}
Based on observation \ref{obs:1}, we propose an approach to preserve the structure of the OCR results. As shown in \Cref{fig:latex element}, we use \latex to encapsulate the OCR text. It is easy for MLLMs to understand \latex code because it usually follows a fixed and easy-to-understand paradigm. The \latex code constrains the content in titles, tables, and figures, providing efficient and additional references when MLLMs answer DocQAs with images. We input the image and the corresponding OCR text into the MLLM to obtain the structured code. We prompt the MLLM to preserve the structure and text related to charts in the image as much as possible. If a figure in the image cannot be converted into \latex content, we instruct the MLLM to use a virtual path to represent the figure while keeping the structure intact. An example is provided in \Cref{fig:latex element}.

We use the proposed method to obtain structured text and compare the performance of MLLMs under three cases: images only, images with OCR text, and images with structured text. The results confirm the hypothesis in \Cref{sec:3.1}, leading to the following observation:
\begin{figure}[!tb]
	\centering
	\includegraphics[width=1.0\linewidth]{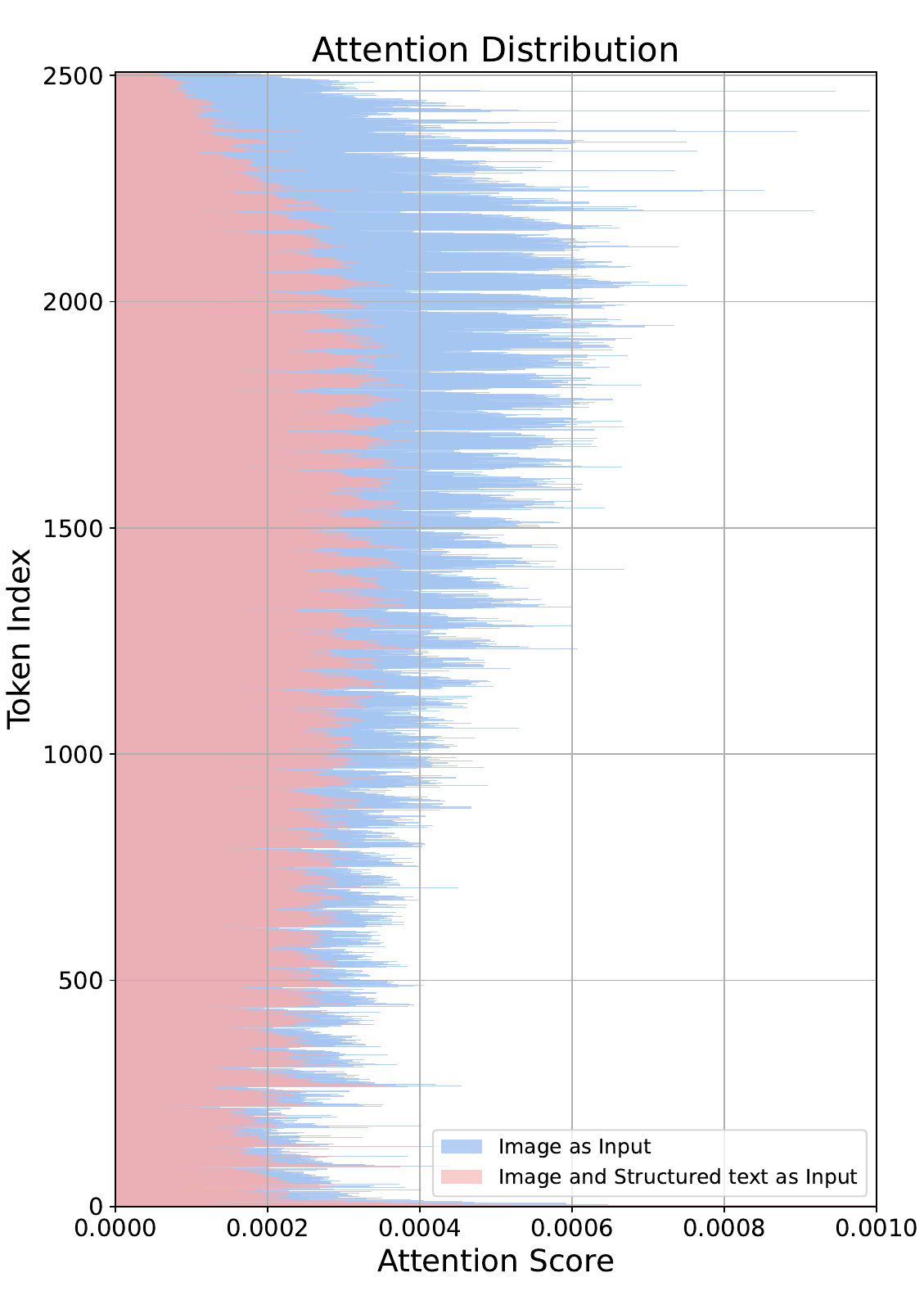}
	\caption{MLLMs are less sensitive to image border tokens when constrained by structured text. The attention distribution shows that attention scores at the borders are lower in the presence of structured text.}	
    \label{fig:attention distribution}
\end{figure}

\begin{observation}\label{obs:2}
Structured text and images work together to improve MLLMs' performance in answering DocQAs.
\end{observation}

Observation \ref{obs:2} shows that the performance of MLLMs are improved significantly simply by preserving the structure of the input text, without the need to provide additional fine-grained positional information to help answer the questions. This observation proves the importance of structure, which means that merely informing the MLLMs of the structure of document helps them gain a better overview and answer questions correctly.

\subsection{Attention transformation with structured OCR text}\label{sec:3.3}
Given observation \ref{obs:2}, this subsection seeks to uncover the underlying reasons why structured textual content matters by analyzing the attention distribution when MLLMs answer questions using only images versus using structured text as references. \Cref{fig:attention distribution} shows that in the structured case, the attention scores are less sensitive to the boundaries of the image and more concentrated on the main body, indicating that the MLLMs know where to focus under the constraints of structured text. Based on these findings and experimental results, we have the following observation:

\begin{observation}\label{obs:3}
Structured text brings structured attention to both texts and images, which directly improves the abilities of MLLMs. This shows that structured attention is the key to helping MLLMs answer DocQAs.
\end{observation}


Observation \ref{obs:3} demonstrates the necessity of structure in document understanding. As shown in the example in \Cref{fig:attention example}, structured text constrains MLLMs and reduces irrelevant attention when focusing on images. This helps MLLMs recover distracted attention and focus on images and relevant text to answer the given question.

\begin{figure*}[!tb]
	\centering
	\includegraphics[width=1.0\linewidth]{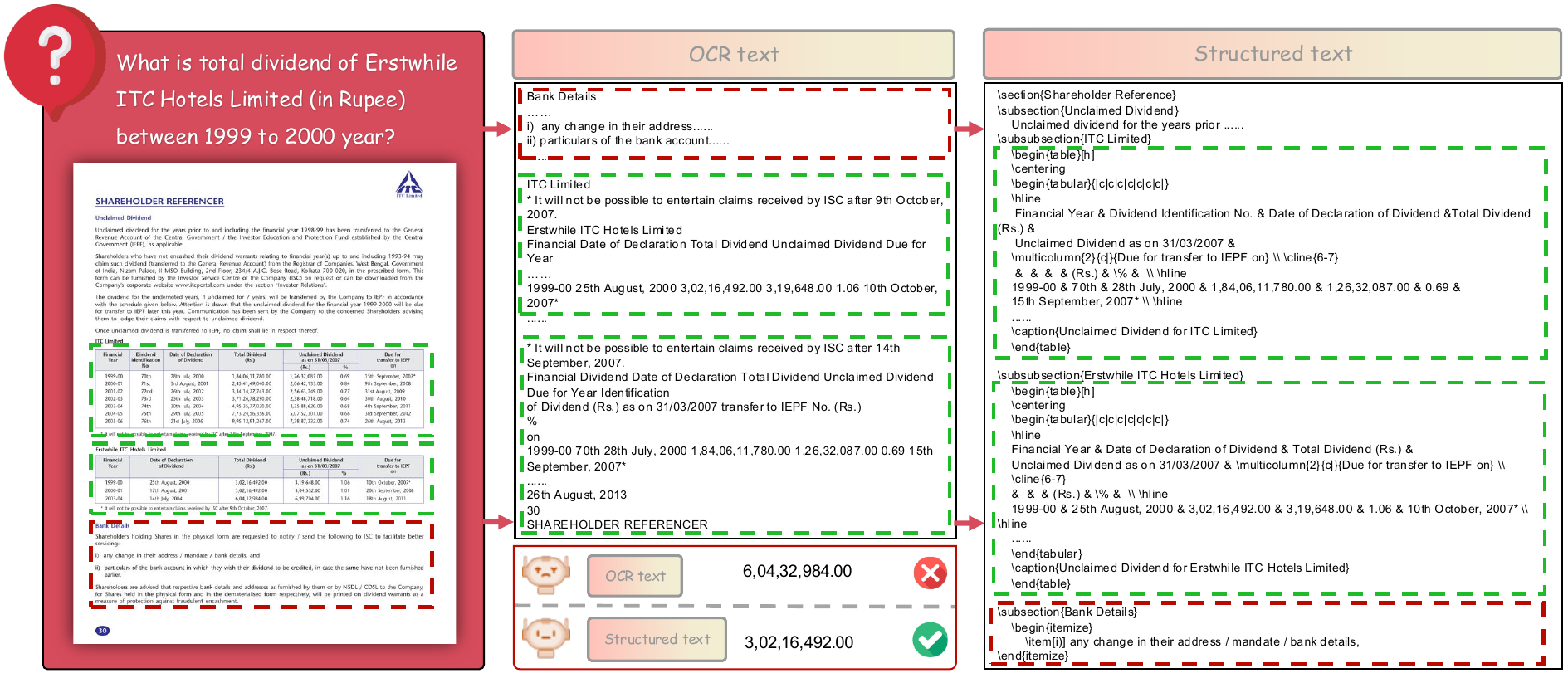}
	\caption{
    A comparison of generated answers from MLLMs using OCR text versus structured text shows that structured text improves MLLM performance on DocQA tasks. The \latex paradigm helps preserve the image’s structure, aiding the model in locating evidence relevant to the question.
}	\label{fig:latex example}
\end{figure*}

\section{Experiments} \label{sec:exp}
We conduct evaluations on four document understanding benchmarks covering multiple scenarios to provide solid evidence supporting the observations presented in this paper.

\subsection{Datasets and models}
\label{sec:data_model}

\paragraph{Implementation Details.} We apply the document preprocessing method from \cite{han2025mdocagent} to obtain textual content extracted using a combination of Optical Character Recognition (OCR) and PDF parsing techniques, and visual content by transforming pages in long documents into images. We locate the evidence pages and apply OCR to the corresponding images to obtain OCR results, then use the approach presented in \Cref{sec:3.2} to obtain the structured text. All experiments are conducted on 4 NVIDIA A100 GPUs.

\paragraph{Models.} Our evaluation uses the following four multimodal LLMs: \qwenA, \qwenB, \llamaM and \PhiV. These models accept both images and text as input and generate answers and analyses for the given questions.

\paragraph{Datasets.} The benchmarks include MMLongBench \cite{ma2024mmlongbench}, LongDocUrl \cite{deng2024longdocurl}, PaperTab \cite{hui2024uda}, and FetaTab \cite{hui2024uda}. These evaluation datasets cover a variety of scenarios, including open- and closed-domain, textual and visual, as well as long and short documents, ensuring fairness and completeness in the evaluation. 

\paragraph{Metrics.} For all benchmarks, we leverage \llamaT as the evaluation model to assess the consistency between the model’s output and the reference answer producing a binary decision (correct / incorrect). We report the average accuracy for each benchmark.
\vspace{-5pt}

\begin{figure*}[!tb]
        \centering
        \includegraphics[width=1.0\textwidth]{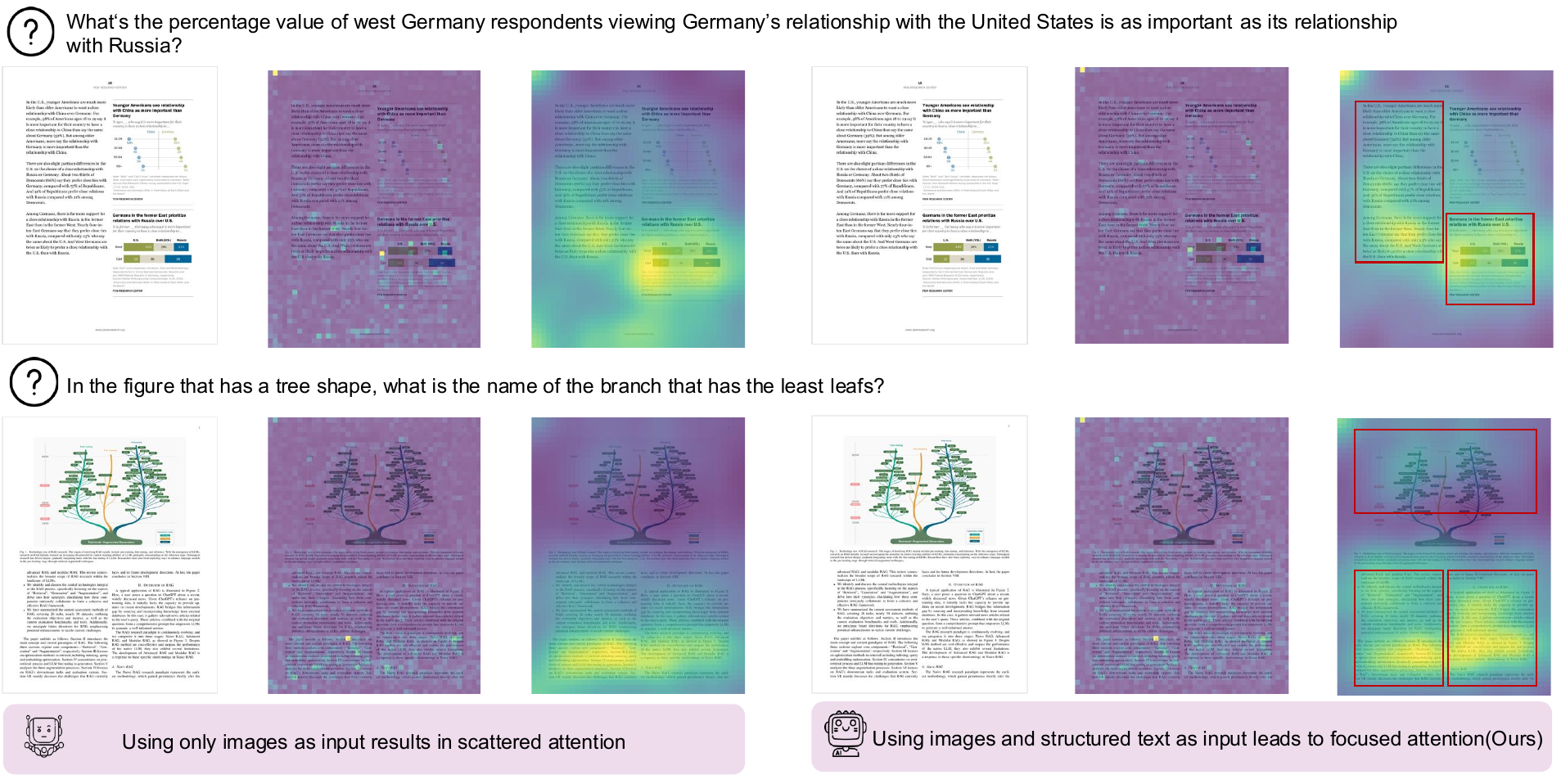}
	\caption{
           An example of attention transformation under two conditions: images alone versus images combined with structured text. The MLLMs tend to focus on text blocks, figures, and charts, rather than the overall layout.
		\label{fig:attention example}}

\end{figure*}


\subsection{Impact of OCR texts} 
\label{sec:imparct of ocr text}
We first compare the performance of MLLMs when using images alone versus using images with the help of OCR text. According to \Cref{tab:exp}, the results show that OCR text impairs MLLMs' performance on MMLongBench, which contains multi-element evidence images including charts, tables, blocks of text, and figures. The additional text information decreases the accuracy of the large model on this dataset from 0.389 to 0.370. The accuracy of MLLMs on other datasets might increase, but this can also impair their ability in specific cases. Providing unstructured OCR text alongside document images sometimes degrades MLLM performance across benchmark datasets, despite increasing the total information available to the model. These results and the case shown in \Cref{fig:latex example} clearly and convincingly support observation \ref{obs:1}, showing that repeated and redundant text information does not help large models answer questions about long documents.

\begin{table*}[t]
	\centering
	\begin{adjustbox}{width=1.0\linewidth}
        \setlength{\tabcolsep}{5pt}
		\begin{tabular}{@{}l|cccccc@{}}
			\toprule
			\textbf{Model}
			& \textbf{MMLongBench}
                & \textbf{LongDocUrl}
                & \textbf{PaperTab}
                & \textbf{FetaTab}
                & \textbf{Avg}
    \\
			\midrule
			\midrule
    \multicolumn{6}{c}{\cellcolor{orange!10}\textbf{Prior Multimodal LLMs}}
    \\ 
    \midrule
    \qwenA w/ image \cite{wang2024qwen2} & $0.292$ & $0.195$ & 0.160 & 0.441& 0.272\\
    \qwenA w/ image+text \cite{han2025mdocagent} & $0.287$ & $0.219$ & 0.188 & 0.507 &0.300\\
    \midrule 
    \qwenB w/ image \cite{qwen2.5-VL}& $0.389$ & $0.197$ & 0.163 & 0.508 & 0.314 \\
    \qwenB w/ image+text \cite{han2025mdocagent} & $0.375$ & 0.399 & 0.229 & 0.557 &0.427\\
    \midrule
   \llamaM w/ image \cite{liu2024improved} & 0.131 & 0.126 & 0.051 & 0.154 & 0.116 \\
    \llamaM w/ image+text \cite{han2025mdocagent} &  0.183 & 0.159 &  0.127 &  0.406 &  0.219 \\
     
    \midrule
    \PhiV w/ image \cite{abdin2024phi} & $0.189$ & $0.131$ & 0.077 & 0.245& 0.161 \\
    \PhiV w/ image+text \cite{han2025mdocagent} & $0.226$ & 0.213 & 0.160 & 0.443 &0.261\\
    
    \midrule
    \multicolumn{6}{c}{\cellcolor{green!10}\textbf{Prior Multimodal LLMs + Structured text (Ours)}}
    \\ 
    \qwenA w/ image & \textbf{0.306} & \textbf{0.229}& \textbf{0.209} & \textbf{0.509} & \textbf{0.313} \\
    + \textbf{ Structured Text} &  \textcolor{green!60!black}{+4.8 \%} & \textcolor{green!60!black}{+17.4\%} &  \textcolor{green!60!black}{+30.6\%} & \textcolor{green!60!black}{+15.4 \%}&\\
    \qwenB w/ image & \textbf{0.435} & 0.221 & \textbf{0.252} & \textbf{0.575} & 0.371 \\
    + \textbf{ Structured Text}  &\textcolor{green!60!black}{+11.8\%} & \textcolor{green!60!black}{+12.2\%} &   \textcolor{green!60!black}{+54.6\%} & \textcolor{green!60!black}{+13.2\%} & \\
    \llamaM w/ image  &  \textbf{0.224}  &  0.153 &  0.122 &  0.388 &  \textbf{0.222} \\
    + \textbf{ Structured Text}  &  \textcolor{green!60!black}{+71.0\%} & \textcolor{green!60!black}{+21.4\%} & \textcolor{green!60!black}{ +139.2\%} &  \textcolor{green!60!black}{+151.9 \% }&\\
     \PhiV w/ image &  \textbf{0.284} &  0.211 &  \textbf{0.224} &  0.429 &  \textbf{0.287} \\
     + \textbf{ Structured Text}  & \textcolor{green!60!black}{+50.3 \% }& \textcolor{green!60!black}{+61.1\%} &  \textcolor{green!60!black}{+190.9\% }& \textcolor{green!60!black}{+75.1\%} &\\
    \midrule
			\bottomrule
		\end{tabular}
	\end{adjustbox}
	\caption{Our proposed structure-preserving method effectively retains the document structure and significantly improves the accuracy of DocQA responses. For the open-domain dataset, we followed the retrieval method used in \cite{han2025mdocagent}, using the top-1 retrieval result as input. On the closed-domain dataset, we used the evidence pages of the dataset as input to test the importance of document structure.}
    \label{tab:exp}
    	\vspace{0mm}
\end{table*}

\begin{table}[t]
	\centering
	\begin{adjustbox}{width=1.0\linewidth}
        \setlength{\tabcolsep}{5pt}
		\begin{tabular}{@{}l|ccc@{}}
			\toprule
			\textbf{Model}
			& \textbf{MMLongBench}
                & \textbf{LongDocUrl}
                & \textbf{Avg}
    \\
			\midrule
			\midrule
    \qwenA  & \multirow{2}{*}{\textbf{0.291}} &\multirow{2}{*}{\textbf{0.175}} &\multirow{2}{*}{\textbf{0.233}} \\
    w/ text  & & & \\
    \qwenA  & \multirow{2}{*}{0.212} &\multirow{2}{*}{0.163} &\multirow{2}{*}{0.188} \\
    w/ structured text  & & & \\
    \midrule
     \qwenB  & \multirow{2}{*}{\textbf{0.318}} &\multirow{2}{*}{\textbf{0.188}} &\multirow{2}{*}{\textbf{0.253}} \\
    w/ text  & & & \\
    \qwenB  & \multirow{2}{*}{0.267} &\multirow{2}{*}{0.155} &\multirow{2}{*}{0.211} \\
    w/ structured text  & & & \\
    \midrule
    \llamaM  & \multirow{2}{*}{\textbf{0.287}} &\multirow{2}{*}{\textbf{0.162}}&\multirow{2}{*}{\textbf{0.225}} \\
    w/ text  & & & \\
     \llamaM  & \multirow{2}{*}{0.246} &\multirow{2}{*}{0.151} &\multirow{2}{*}{0.199} \\
    w/ structured text  & & & \\
    \midrule
     \PhiV  & \multirow{2}{*}{\textbf{0.299}} &\multirow{2}{*}{\textbf{0.185}} &\multirow{2}{*}{\textbf{0.242}} \\
    w/ text  & & & \\
     \PhiV & \multirow{2}{*}{0.244} &\multirow{2}{*}{0.167} &\multirow{2}{*}{0.206} \\
    w/ structured text  & & & \\
    \midrule
			\bottomrule
		\end{tabular}
	\end{adjustbox}
	\caption{Performance comparison on different datasets. The ablation results show the importance of combining structured text and image. When structured text is combined with an image, it results in structured attention, which eventually helps MLLMs answer questions.}
    \label{tab:ablition text}
    	\vspace{0mm}
\end{table}

\begin{table*}[t]
	\centering
	\begin{adjustbox}{width=1.0\linewidth}
        \setlength{\tabcolsep}{5pt}
		\begin{tabular}{@{}l|ccccccc@{}}
			\toprule
			\textbf{Model}
			& \textbf{Chart}
                & \textbf{Table}
                & \textbf{Pure-Text}
                & \textbf{Generalized-text}
                & \textbf{Figure}
                & \textbf{Avg}
    \\
			\midrule
			\midrule
    \multicolumn{7}{c}{\cellcolor{orange!10}\textbf{Prior Multimodal LLMs}} \\
    \qwenA w/ image & 0.278 & 0.304 & 0.311 & 0.311& 0.318 & 0.304\\
     \qwenA w/ image+text & 0.244 &0.300 & 0.358 & 0.294 & 0.301 & 0.299\\
    \midrule
    \qwenB w/ image & 0.392 & 0.350  & 0.427 & 0.387 & 0.418& 0.395\\
    \qwenB w/ image+text & 0.375 & 0.367 & 0.374 & 0.420 &0.388 & 0.285\\
    \midrule
    \llamaM w/ image & 0.120 & 0.060  & 0.116&0.126 &0.134 & 0.111\\
    \llamaM w/ image+text & 0.114 & 0.106 & 0.195& 0.151& 0.154 & 0.144\\
    \midrule
    \PhiV w/ image & 0.176 & 0.147  & 0.185 & 0.160 & 0.244 & 0.182\\
    \PhiV w/ image+text & 0.205 & 0.203 &0.291& 0.261& 0.224& 0.237\\
    \midrule
    \multicolumn{7}{c}{\cellcolor{green!10}\textbf{Prior Multimodal LLMs + Structured text (Ours)}} \\
    
    \textbf{\qwenA w/ structured text} &  \textbf{0.307} & 0.267  &  0.348 & 0.261 &0.288 &0.294\\ 
    \textbf{\qwenB w/ structured text} &  0.364 &  0.364  &  0.374 & 0.328& 0.385&0.363 \\
    \textbf{\llamaM w/ structured text} &  \textbf{0.159} &  \textbf{0.129}  &  \textbf{0.205} &\textbf{0.160} & \textbf{0.161} &\textbf{0.163} \\
    \textbf{\PhiV w/ structured text} &  \textbf{0.261} &  \textbf{0.263} &  \textbf{0.315} & 0.227 & 0.231&\textbf{0.259}\\
    \midrule
			\bottomrule
		\end{tabular}
	\end{adjustbox}
	\caption{Performance comparison across different evidence sources on MMLongBench reveals how various document elements demonstrate the differing effects of structured input on document understanding.}
    \label{tab:ablition element}
    	\vspace{0mm}
\end{table*}

\subsection{Performance improvements with structured text} 
\label{sec:performance structed text}
We consider the impact of structure on the process of MLLMs understanding documents. We compare using structured text as a reference with using only images for understanding. On MMLongBench, adding structured text information increases the accuracy of \qwenB from 0.389 to 0.435. This 10\% performance improvement fully demonstrates the importance of structure in long document reading. The structure of different elements in the evidence images is preserved as much as possible due to the instructions given to the MLLM. The structured text surprisingly enhances the ability to answer questions. On MMlongBench, every models' performance are improved because almost all evidence pages in this dataset contain structured tables and graphs, which can be fully interpreted by the \latex paradigm combined with OCR text. Additionally, the accuracy of \qwenB is significantly higher compared to \qwenA, demonstrating that \qwenB has a stronger ability to understand \latex and make better use of structure.

\subsection{Attention Analysis} 
\label{sec:attention analysis}
We follow \cite{zhang2024cls} to conduct attention analysis in order to understand why structured text matters to MLLMs. We analyze attention from two perspectives: attention distribution and attention transformation across different cases. We present the following findings to support Observation \ref{obs:3}.

\paragraph{Attention Distribution.} We consider 370 samples with the same number of image tokens to demonstrate the distribution of attention transformation with or without structured text as references when MLLMs answer questions based on evidence images. We use \qwenB as the main model to generate answers on MMLongBench. In the case where MLLMs rely only on images, the attention distribution shows that the model is sensitive to boundary tokens of images and exhibits an uneven distribution of attention. These results indicate that MLLMs treat almost every image token equally and have no clear focus on specific regions of the original images. In contrast, \Cref{fig:attention distribution} shows that MLLMs are constrained when using structured text as references. MLLMs learn to focus on useful image tokens located in figures, blocks of text, or tables. This transformation of attention distribution highlights the importance of structured text.

\paragraph{Case Study.} We perform a case study to better understand attention transformation with the help of structured text. \Cref{fig:latex example} illustrates an example. The question requires MLLMs to find the percentage of West Germany respondents who view Germany's relationship with the United States as equally important as its relationship with Russia, based on the chart and corresponding text. The attention map in \Cref{fig:attention example} shows that attention is distributed everywhere without the control of structured attention, even in the blank areas of the given image. With the constraint of structured attention, attention focuses on blocks of text, charts, and so on. We conclude that structured text helps MLLMs reduce attention loss, guides them where to look, and increases the probability of finding fine-grained evidence regions. All these constraints enable MLLMs to better understand documents and answer DocQAs more effectively.
\begin{figure}[!tb]
	\centering
	\includegraphics[width=1.0\linewidth]{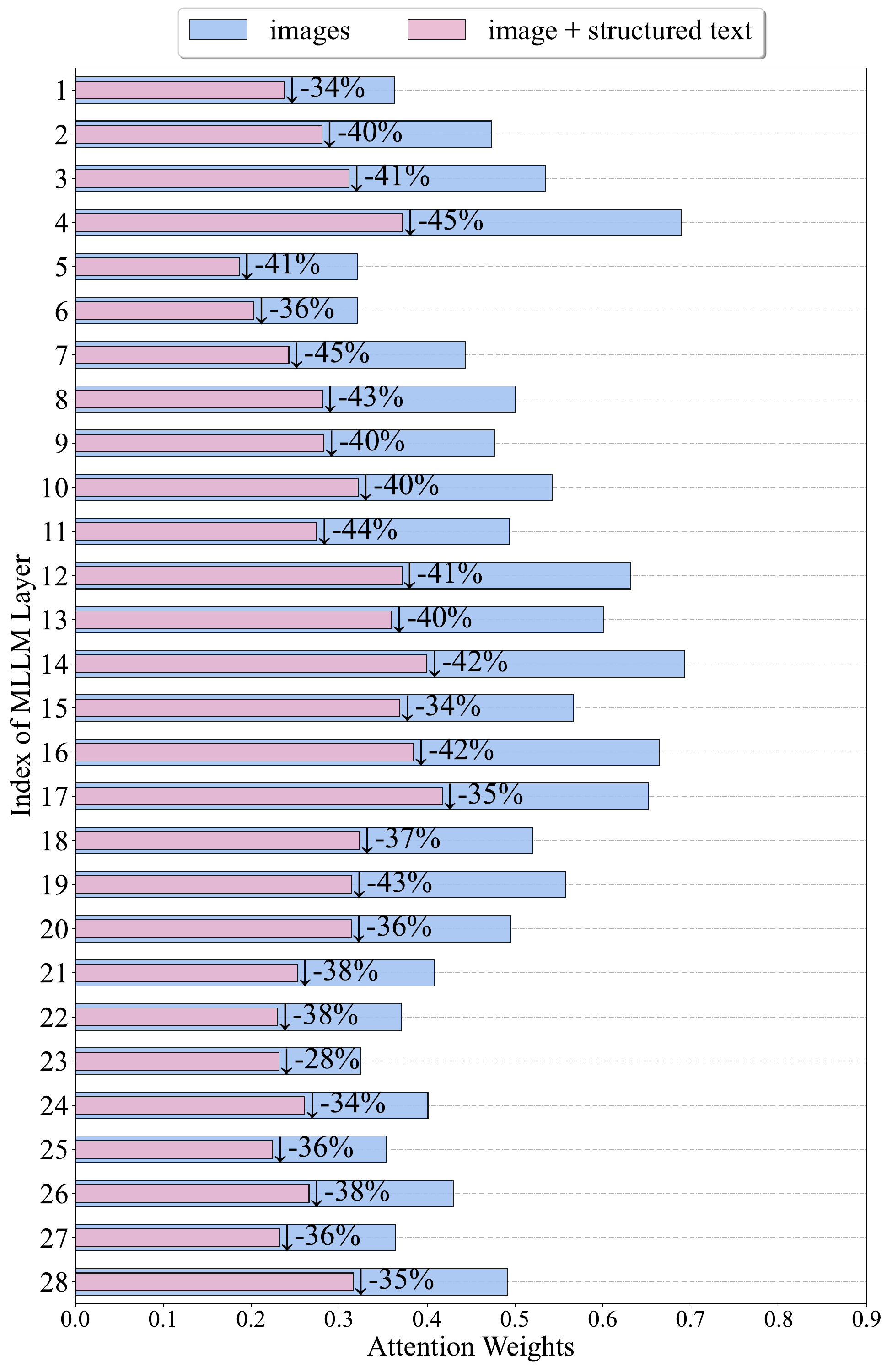}
	\caption{
	Comparison of attention weights across different layers of MLLMs for images alone versus images combined with structured text.
		\label{fig:attention layer}}
\end{figure}
\subsection{Ablation Studies}\label{sec:ablation studies}
We almost obtain the best results on almost every dataset and model by using structured text and images as input. We further conduct experiments on the following cases: structured text as input, OCR text as input, and the \latex format acting as a placeholder without specific text information. Based on these cases, we aim to demonstrate the necessity of combining structured text and images. \Cref{tab:ablition text} shows that cases relying only on OCR text or only on structured text result in poor performance of MLLMs. Structured text is ineffective when not input alongside images.

We conducted ablation experiments categorized by different evidence elements from MMLongBench. The results in \Cref{tab:ablition element} indicate that for questions in documents that necessitate answers from charts or tables, using structured input provides a more pronounced performance gain than for other types of document elements. This observation further underscores the critical role of structured input.In the cases needed images to answer questions, the accuracy is lower. This is understandable because the lack of images causes information loss in cases where MLLMs need figures from the evidence images to answer questions. The accuracy with structured text ans images is better than with OCR text and images, further highlighting the importance of structure. This experimental result convincingly demonstrates the importance of structure. We conclude that the combination of structured text and images is key to improving the ability of MLLMs in long document understanding, supporting Observation \ref{obs:3} from an ablation perspective.

\section{Conclusion}
Preserving the structure of the input to improve the ability of general-purpose multimodal MLLMs is essential to comprehend the underlying patterns and key factors in document understanding. 
This work makes a step in this line. 
We propose to use \latex paradigm to keep the structure of OCR text, the images combined with structured text efficiently improve the accuracy of answering questions in document understanding. Furthermore, we analyze the attention transformation of different kinds of input. This study find the structured attention is the key to make MLLMs understand better in answering DocQAs after comparing other inputs. 
Future work includes proposing novel and efficient structure extraction or attention control method to effectively unlock the ability of general-purpose MLLMs in document understanding.

\section{Limitations}
The structure-preserving method proposed in this paper is an early attempt to enhance the document reading capability of MLLMs.
Could more advanced and generalized structure-preserving methods be adopted, or should we design plugins from the perspective of attention in MLLMs themselves to ensure structural attention during document understanding?
We hope that future work will explore these issues and further improve document comprehension, enabling it to serve more practical applications.




\bibliography{acl2020}


\newpage

\end{document}